\pdfoutput=1

\documentclass[twocolumn]{article}
\usepackage{times}
\usepackage{soul}
\usepackage{url}
\usepackage[utf8]{inputenc}
\usepackage{graphicx}
\usepackage{amsmath}
\usepackage{amsthm}
\usepackage{booktabs}
\usepackage{algorithm}
\usepackage{algorithmic}
\usepackage{multicol}
\usepackage{multirow}
\usepackage{float}

\usepackage[font=normalsize]{caption}
\usepackage{hyperref}
\hypersetup{
     colorlinks = true,
     linkcolor = blue,
     anchorcolor = blue,
     citecolor = cyan,
     filecolor = blue,
     urlcolor = blue
     }

\usepackage{microtype}
\usepackage{subfigure}
\usepackage{adjustbox}
\usepackage{geometry}

\makeatletter
\newcommand{\printfnsymbol}[1]{%
  \textsuperscript{\@fnsymbol{#1}}%
}
\makeatother

\title{
    Memory Defense: More Robust Classification via a Memory-Masking Autoencoder
}

\author{
    Eashan Adhikarla\thanks{Equal Contribution}  , Dan Luo\printfnsymbol{1}, Brian D. Davison \\\\
    Department of Computer Science and Engineering, \\
    Lehigh University, USA \\
    \{eaa418, dal417, bdd3\}@lehigh.edu
}

\date{}
\newgeometry{vmargin={20mm,25.4mm}, hmargin={25.4mm,25.4mm}}   

\begin{document}
\maketitle

\begin{abstract}
    Many deep neural networks are susceptible to minute perturbations of images that have been carefully crafted to cause misclassification. Ideally, a robust classifier would be immune to small variations in input images, and a number of defensive approaches have been created as a result. One method would be to discern a latent representation which could ignore small changes to the input. However, typical autoencoders easily mingle inter-class latent representations when there are strong similarities between classes, making it harder for a decoder to accurately project the image back to the original high-dimensional space. We propose a novel framework, Memory Defense, an augmented classifier with a memory-masking autoencoder to counter this challenge. By masking other classes, the autoencoder learns class-specific independent latent representations. We test the model's robustness against four widely used attacks. Experiments on the Fashion-MNIST \& CIFAR-10 datasets demonstrate the superiority of our model. We make available our source code at GitHub repository: \href{https://github.com/eashanadhikarla/MemoryDef}{https://github.com/eashanadhikarla/MemoryDef}
\end{abstract}

\section{Introduction}
\label{submission}

    Deep learning (DL) has proved to be an incredibly powerful tool. With advances in deep learning, organizations are reshaping their businesses. For better performance, organizations are deploying Deep Neural Network (DNN) models that enable systems to disentangle and learn complex concepts, thereby providing a better way to solve classification tasks. However, organizations that completely rely on deep learning applications become more prone to security threats. With advancements in image classification, intelligent systems can often be easily misled by marginally different input images \cite{kurakin2016adversarial,goodfellow2014explaining,CarliniW16a}. If a small perturbation can drastically change the model behaviour then the model is not very robust, as it is not learning the underlying concept of the image (for image classification tasks). 
    While deploying DL models, a system should not only perform well but it should be reliable and robust against attacks. In recent years, adversarial robustness has been an open area of research. Learning the nature of the dataset and being able to recognize adversarial attacks supports not only the model's generalization but also makes the model adversarially robust \cite{Stutz2019CVPR}. 
    A good generalized model will intelligently learn the critical underlying concepts of the dataset and also utilize them to disentangle the spatial features of the data and can fix the foreign-looking malicious data during reconstruction. Designing robust DNN models is a crucial step toward a trustworthy AI system and a stronger defense towards attacks.
    \begin{figure}
        \center
        \includegraphics[scale=0.36]{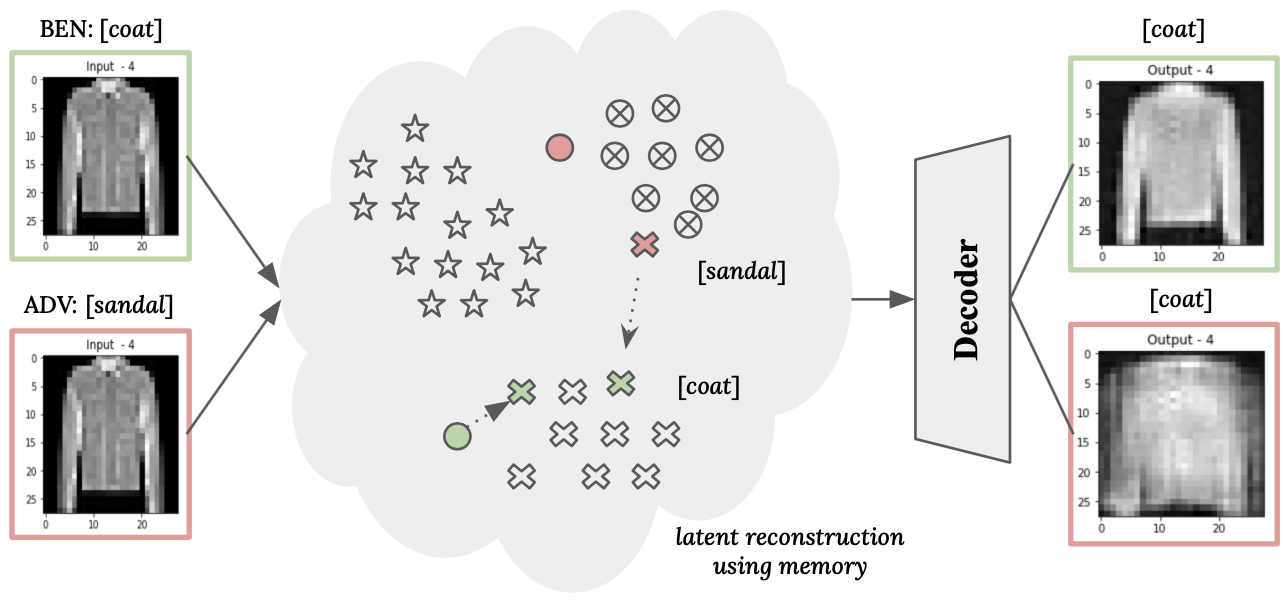}
        \caption{\textit{Two input images with their predicted labels are encoded into the feature space, where green represents benign and red represents an adversarial example. The encoding is reconstructed with useful memory features of ``coat" class and classified correctly as ``coat".}
        }
        \label{fig:figure1}
    \end{figure}

    Due to the growth in applications of DL, adversarial attacks and defenses have become increasingly popular. Researchers have developed numerous defense mechanisms in recent years. Deep learning applications have attracted significant attention toward unsupervised learning approaches for anomalous and adversarial example detection based on their noteworthy performance on benchmark tasks. Autoencoders are an unsupervised learning tool that is commonly used for the task of representation learning and have also gained popularity in deep learning for anomaly detection \cite{10.1145/3097983.3098052} and adversarial example detection \cite{8411207}. Recent advances use deep convolutional autoencoders to learn extracted features from the latent space for binary and multi-class classification. Our approach using memory-based reconstruction is influenced significantly by the Memory Augmented Autoencoder (MemAE) \cite{gong2019memorizing}. We incorporate a deep autoencoder with a masked attention module to extract the useful features from the latent dimension for image reconstruction to be able to differentiate between benign and adversarial images. Figure \ref{fig:figure1} shows an example for a benign and adversarial example from fashion-MNIST dataset passing through our defense model. The key contributions that we make in this paper can be summarized as follows: 
    \begin{itemize}
        \item 
        We propose a unique memory-masking autoencoder strategy that learns to correctly map the latent representations in the feature space for feature extraction.
        \item
        To restrict the inter-manifold learning of different classes, we propose a masking strategy over the attention weight vectors that are used for reconstructing the image.
        \item
        To effectively classify adversarial examples, we define the latent representation task as a classification problem by jointly training the autoencoder with the classifier. Experiments show Memory-Defense outperforms multiple state-of-the-art methods.
        \item
        We evaluate Memory Defense and study how the different parts of the module enhance the overall defense system.
    \end{itemize}

\section{Background}
    \subsection{Adversarial Attacks}
        Adversarial attacks became widely popular after discovering that deep neural networks are highly discontinuous in mapping the inputs to outputs \cite{42503}. This can easily lead to errors in a classification model if a perturbation successfully changes the mapping. The adversarial perturbation can be applied in multiple ways to attack the DNN model. One way is to spread out the minimal perturbation across the entire range of pixels. Recent studies have also shown numerous other ways such as: attacking a small patch of image with perturbation can also fool the DNN model easily \cite{Sharif16AdvML}, using generative adversarial network (GAN) based attack where the generator learns to find an optimal adversarial noise \cite{ijcai2018-543}. An adversarial attack's overall goal is to cause the target model to generate incorrect outputs. The differences caused by adversarial changes are intended to be smaller than that observable by the human eye. For an adversarial attack to be successful, the adversarial image $x_{adv}$ should fall close to the benign image's ($x$) latent space at a certain distance. The distance is measured using $L_{p}$ norm distance metric that denotes the similarity between the image. The distance between any benign and adversarial image is denoted by $||x-x_{adv}||_{p}$ where $L_{0}, L_{2},$ and $L_{\infty}$ are types of distances. The general norm function is described  \cite{magnet} as:
        \begin{align}
            ||d||_{p} = \left( \sum_{i=1}^{n}|d_i|^p\right)^\frac{1}{p}
            \label{eq:norm}
        \end{align}
        where $i$ is the dimension of the $d$ vector and $p$ is the norm. Most researchers do not consider $L_{0}$ a norm, because mathematically it returns the non-zero elements from the image vector. The $L_2$ norm measures the short distance matrix, popularly known as the Euclidean distance matrix. It calculates the square root of the sum of squared distances between any two points, i.e., it returns the hypotenuse distance between the two points. Lastly, $L_\infty$ returns the absolute maximum value out of all the sets of values. We evaluate our model's defensibility against the following four popular adversarial attack algorithms.
    
    \paragraph{Fast Gradient Sign Method (FGSM).}
        FGSM \cite{goodfellow2014explaining} makes a single-step gradient update in the direction of the model's loss function $J(x, y,\theta)$, i.e., with respect to the input feature vector, where $x$ represents the input example and $y$ represents the label. The gradient is updated with an intensity of $\epsilon$ over the loss $J$.
        \begin{align}
            x_{adv} = x + \epsilon*sign(\nabla_{x}J(\theta, x, y))
            \label{eq:fgsm}
        \end{align}
    \paragraph{Basic Iterative Method (BIM).}
        The basic iterative method \cite{kurakin2016adversarial} calculates the step size multiple times (unlike FGSM) and clips the results in each step ensuring they do not deviate much from the original image. Otherwise, this is the same form as Equation \ref{eq:fgsm}, where $t$ represents the iteration number.
        \begin{align}
            x_{adv_{t+1}} = clip [x_t + \epsilon*sign(\nabla_{x_t}J(\theta,x_t,y))]
            \label{eq:bim}
        \end{align}
    \begin{figure*}
        \center
        \includegraphics[scale=0.39]{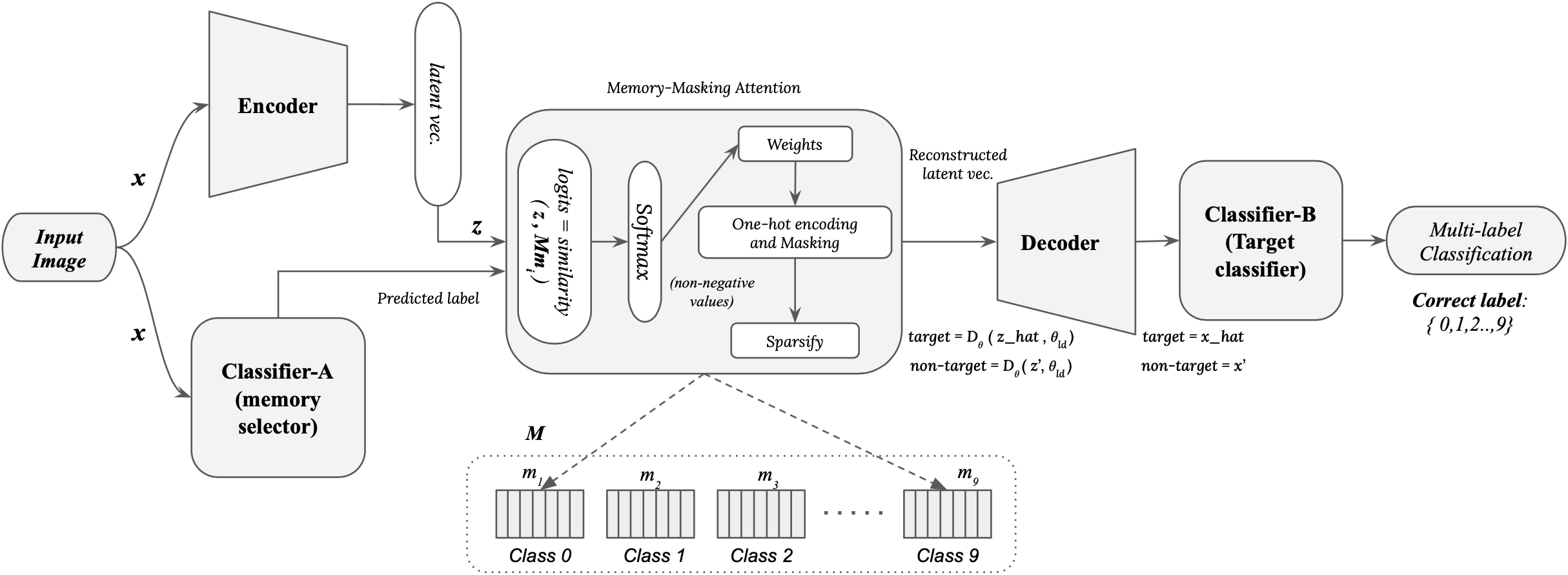}
        \caption{\textit{Presenting the workflow of the entire Memory Defense system. The image passes through the Memory Augmented Autoencoder with an aim to be reconstructed with the least possible error through the right memory slot. The target model predicts the reconstructed image label based on reconstructed image.}}
        \label{fig:figure2}
    \end{figure*}
    \paragraph{Projected Gradient Descent (PGD).}
        In the adversarial attack literature, projected gradient descent \cite{madry2018towards} (variant of FGSM) calculates the gradient descent with respect to the inner maximization ($\delta$) to maximize the objective, ensuring that $\delta$ remains in the norm bound. In Equation [\ref{eq:pgd}], $\rho$ means the norm of interest for projection.
        \begin{align}
            \hat{x_{adv_{t+1}}} = \rho [x_t + \epsilon*sign(\nabla_{x_t}J(\theta,x_t,y))]
            \label{eq:pgd}
        \end{align}
    \paragraph{Carlini \& Wagner (CW).}
        One of the strongest attacks introduced by Carlini and Wagner \cite{CarliniW16a} is an attack which surpasses the problem of defensive distillation, i.e., training to quantify the predictability of another model that was trained earlier. The $\kappa$ parameter stimulates to misclassify with confidence.  There are several versions of the attack making use of three different distance metrics; $l_{0}$, $l_{2}$, and $l_{\infty}$. To evaluate our approach we use the $l_{2}$ distance metric for attack generation.
        \begin{align}
            f(x_{adv})=\text{max}(\text{max}\{z(x_{adv})_i:i\neq t\}-z(x_{{adv}_t}, -\kappa))
            \label{sec:cw}
        \end{align}
        Attacks generated by all the methods described in this section use a perturbation scheme ($\delta$) that is added to the benign input array ($X$) (Equation \ref{eq:6}). In particular we use the $l_{\infty}$ norm for FGSM, BIM \& PGD and $l_{2}$ for CW attack. Our proposed approach is to retrieve and learn the underlying class relevant memory contents which will purely reconstruct the example with important memory features of the respective genuine class.
        \begin{align}
            x_{adv} = x + \delta
            \label{eq:6}
        \end{align}
    \subsection{Defense} 
        Researchers have also come up with a number of defensive strategies. We summarize these strategies to counter attacks into the following three categories. 
        
        \paragraph{Adversarial training.} A training method where the model is exposed to both the normal dataset and the adversarial dataset. Numerous studies have shown to improve the robustness of the system significantly through adversarial training \cite{DBLP:conf/iclr/WongRK20,10.1145/3128572.3140449,Miyato2015DistributionalSW,madry2018towards}, however adversarial training also generally causes adversarial over-fitting \cite{AOF} phenomenon where after a certain point in training the classifier's robustness starts to decrease substantially. The goal of a robust model is to defend the model even when the attacker has complete knowledge of its parameters. One way of achieving robustness is by adding adversarial examples to the training and helping the model to classify such perturbations correctly \cite{Huang2015LearningWA,Shaham2018UnderstandingAT}.
        
        \paragraph{Architecture Engineering.} Another way of building adversarial robustness is by engineering the model architecture to make it less sensitive to attacks and hide gradients w.r.t.\ the input image reducing the adversarial impact \cite{madry2018towards,42503}. a.) Hiding the latent space of one class from another by maximally separating the class-specific prototypes \cite{Mustafa_2019_ICCV} restricting the adversarial attack to enter foreign class latent space. b.) Defensive Distillation uses additional class characteristics like soft labels to train the model, thereby reducing the adversarial attack's sensitivity \cite{7546524}. c.) Shield \cite{10.1145/3219819.3219910} uses an ensemble design architecture \cite{Liu2018TowardsRN,Strauss2017EnsembleMA,He2017AdversarialED} to vaccinate by random compression levels, potentially weakening attacks.
        
        \paragraph{Pre-processing \& Post-processing} This strategy incorporates a filtering process to eliminate adversarial noise from the network \cite{Buckman2018ThermometerEO,Prakash2018ProtectingJI}. Pre-processing can be performed on the dataset (such as data augmentation, transformation, encoders, etc.) and on the input sample to fine-tune the input and make it easier for the model \cite{DBLP:conf/icml/VincentLBM08,DBLP:conf/icufn/CalimbahinPP19}. Our design focuses on the robust multi-class classification to directly filter out the adversarial images' perturbations to make the target classification simple.

\section{Memory Defense}
\subsection{System Overview}
    \paragraph{Components.}
        Our proposed method, \textbf{Memory Defense}, comprises a \textit{memory selector}, a \textit{memory augmented deep autoencoder with masking}, and a \textit{target classifier} as shown in Figure \ref{fig:figure2}. We call the memory selector and the target classifier as \textit{classifier-A} and \textit{classifier-B} respectively for simplicity. The memory augmented autoencoder with masking contains an encoder for dimensionality reduction, a memory module to retrieve the relevant memory contents, and a decoder to reconstruct the input image. Here, we use the deep autoencoder with the same target space as the input space which makes it a special case of encoder-decoder.
    
\begin{figure}
        \center
        \includegraphics[scale=0.25]{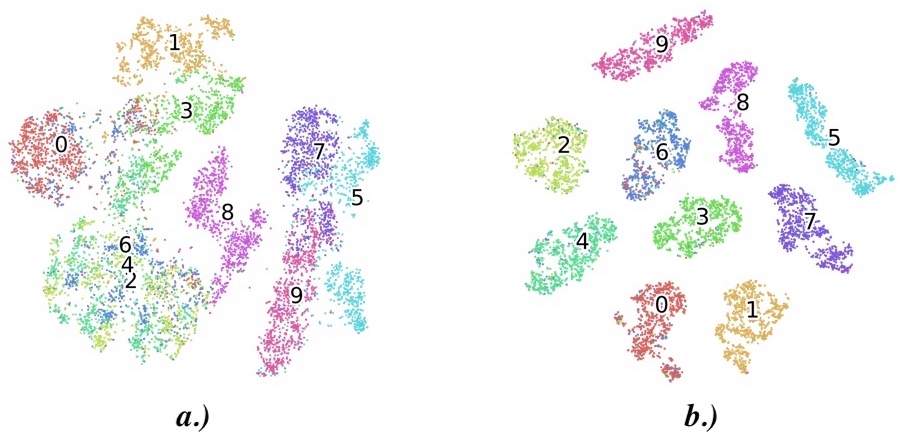}
        \caption{\textit{To identify the structure of data distribution and the generalizability of the model, we plot the latent space of fashion-MNIST dataset without and with memory from the bottleneck layer of the encoder using t-Distributed Stochastic Neighbor Embedding (t-SNE) \cite{vanDerMaaten2008}. Best viewed in color.
        }}
        \label{fig:figure3}
    \end{figure}

    \paragraph{Training.}
        For the training of the overall pipeline, we divide the training in two steps. We first separately train classifier-A (the memory-selector) with equal benign and adversarial examples in the training dataset using PGD attack \cite{madry2018towards}. We perform the training in an offline learning manner, i.e., we generate all the adversarial examples corresponding to classifier-A and then batch them with the training samples for the final training of classifier-A. Secondly, the training for the memory augmented autoencoder and the target classifier uses only benign examples from the dataset. This makes the training of the stack faster, and the model generalizes well in projecting benign examples' prototypes correctly in the latent space as shown in Fig.~\ref{fig:figure3}b. During training of the autoencoder, we minimize a custom loss function (described in Sec.~\ref{sec:loss}) using  back-propagation.
        
        The goal is to optimize the memory slot of each class $m_{i}$ (where M$= \{m_{1}, m_{2}, m_{3}, ..., m_{N}\}$, $m_{i}$ are multiple row-vectors forming a slot, and $N\text{ is the number of classes}$), such that minimization of the reconstruction error of the benign class examples facilitates learning to map to their respective class prototypes in the latent representation. On the other hand, two images that semantically map far away from each other in high-dimensional space could map closely in low-dimensional space. Therefore, we expect the model to produce higher reconstruction errors for anomalous and adversarial examples that will eventually cause the model to drag them to their true class prototypes by the latent reconstruction from the memory attention. Here, one-hot masking over the attention helps restrict inconsistent mappings by utilizing a minimal memory weight of relevant contents of the image to reduce the adversarial impact. Overall, the autoencoder, augmented by memory slots, gives more clarification to adversarial examples by producing higher reconstruction errors that are close to benign prototypes in low-dimensional space.
    
    \paragraph{Workflow.}
        We define a manifold as a proximity region in the low-dimensional space that is entirely different for each class, given that the classes in a dataset are mutually exclusive. Figure \ref{fig:figure3}b. shows the manifolds of each class generalized well over memory augmented autoencoder. Each class is bounded by a boundary known as the manifold of that class. Figure \ref{fig:figure2} shows the entire workflow of the Memory Defense system. The input image $x$ is passed through the encoder and the memory selector simultaneously. The task of the encoder is to generate a low dimension vector out of the higher-dimensional input. The encoder performs dimensionality reduction resulting in a latent space $L(x)$. The input is also passed through classifier-A which generates a prediction of a class with highest probability score for the input image. The attention mechanism computes the attention relevant contents of the class, memory logits using cosine similarity. Logits produce real valued vectors that are transformed into values between 0 and 1, where all the real valued vectors sum to 1. We call this vector as attention weight $\Bar{w}$. Provided the prediction label through classifier-A, we mask the rest of the weights belonging to non-target classes. Using the hardshrinked-masked attention weight, the decoder reconstructs the input example with a low reconstruction error. Here, the reconstruction aims to resolve classifier-A errors in the final prediction by utilizing minimal addressing weights. Now the target classifier is expected to have a noise-free reconstructed image through original class memory contents. This makes the task for target classifier easier for final prediction.

\subsection{Memory Attention with Masking}
    The memory masking is performed by a two-dimensional matrix defined with respect to the memory size $M$ and latent representation $z$ (as in Eq.~\ref{eq:7}) generated by the encoder's bottleneck: $z = E_{\theta}(x; \theta_{l_{e}})$. Let the encoder be represented by a function $E_{\theta}(x; \theta_{l_{e}}): E^{l_{e}}(E^{l_{e}-1}(...E^{2}(E^1(x))))$ where, $l_{e}$ represents the layers in the encoder, $\theta$ represents the parameters involved, and $x$ as the input example. Similarly, let the decoder be represented by a function $D_{\theta}(\hat{z}; \theta_{l_{d}}): D^{l_{d}}(D^{l_{d}-1}(...D^2(D^1(\hat{z}))))$ where, $l_{d}$ represents the layers in the decoder, and $\theta$ represents the parameters involved.

    Weight initialization can be an important step for the memory module to prevent the model from vanishing or exploding the gradient which can leave the model unconverged or could slow the convergence rate. We therefore use the Kaiming-uniform \cite{kaiming} weight initialization technique. 
    Memory masking attention takes paired input of the latent vector of the image and a predicted label corresponding to the highest probability score from classifier A. It then calculates cosine-similarity between the latent representation $z$ and the memory matrix $M$ containing multiple vectors $m_{i}$ corresponding to each class. Cosine-similarity selects only a few relevant contents as memory items as fundamental features for reconstruction based on chosen memory size. Cosine-similarity is a measurement of pair-wise distance between two vectors. Equation~\ref{eq:8} shows $z$ that represents the extracted features from the bottleneck layer of the encoder and $M_{m_{i}}$ as two vectors. The similarity terms $logits_i$ are converted to normalized probability distribution as shown in Equation \ref{eq:9}. $exp(logits_i)$ is applied element-wise on all the vector entries. The denominator is a normalization term where all the entries sum to 1. Generating $\Bar{w}$, a soft addressing vector, in Equation \ref{eq:9} through softmax function gives the entire attention weight of the memory $M$. To restrict the inter-manifolds from learning different classes we propose one-hot memory masking over the attention weight $\Bar{w}$, ensuring that back-propagation only updates the relevant memory slot.
    \begin{gather}
        \label{eq:7} z = E_{\theta}(x; \theta_{l_{e}}) \\
        \label{eq:8} \text{logits}_{i} = \frac{z\cdot M_{m_{i}}}{max(\lVert z \lVert \cdot \lVert M_{m_{i}} \lVert)} \\
        \label{eq:9} \Bar{w} = \frac{exp(\text{logits}_{i})}{\sum_{j=1}^{N}exp(\text{logits}_{j})}
    \end{gather}
    We use two hyperparameters in the memory masking attention module: number of memory slots $m_{i}$ needed per-class for the latent reconstruction using useful memory items, and the hard-shrinkage parameter $\phi$ for sparsifying the overall vector. Shrinking the soft-addressing vector is another useful step in making the vector less dense. Otherwise, there are more chances of adversarial examples to be reconstructed with small errors as benign, making it harder for the target classifier to classify that final prediction. For instance, if the number of memory slots (weights) $M$ is 100 (such that, $m_{i} \in M$, and $i\in[0,9]$), we allocated 10 weight slots to each class for a dataset with 10 classes. 
    
    In Equation~\ref{eq:10}, the one-hot coding scheme generates a $mask$ and $unmask$ vector both of the same size of the memory vector. The $mask$ matrix contains high bit (1's) on the predicted memory slot and rest all the slots as low bit (0's). 
    \begin{equation}
    \label{eq:10}
        \begin{aligned}
          w = \Bar{w} \cdot \text{mask} \\
          \hat{w} = \Bar{w} \cdot \text{unmask}
        \end{aligned}
    \end{equation}
    Conversely, $unmask$ is the compliment of the mask vector. We compute the masking by dot product of weight matrix with $mask$ and $unmask$ to produce two attention weight vectors.
    \begin{gather}
        \label{eq:11} \hat{z} = w \cdot M \\
        \label{eq:12} \hat{x} = D_{\theta}(\hat{z}; \theta_{l_{d}})
    \end{gather}
    Utilizing the masked memory weight and memory $M$ of fixed dimension, we multiply (\ref{eq:10}) to retrieve the latent reconstruction $\hat{z}$ of the same shape as $z$. Now, the new projection $\hat{z}$ is fed into the decoder for the reconstruction of the image. More details can be found in Sec.~\ref{sec:loss}. 
    
    \begin{table*}
        \caption{\label{tab:fmnist} Model Performance - \textbf{Fashion-MNIST} ($L_\infty$ = 9/25), where pixel values are clipped in [0,255]. ``ben'' \& ``pdg'' in the subscript are benign and pdg-adversarial training respectively}
        \vspace{-0.5cm}
        \begin{center}
            \begin{tabular}{c c c c c c}
            \midrule
            \textbf{Model} & CLEAN & FGSM & BIM & PGD & CW\\
            \midrule
            \midrule
            $\text{ResNet-50}_{ben}$ & \textbf{93.66} & 62.83 / 62.52 & 59.88 / 56.10 & 59.06 / 59.08 & 48.61 / 45\\ 
            
            PixelDefend  & 89 & 87 / 82 & 85 / 83 & NA / NA & 88 / 88\\ 
            
            ComDefend & 93 & 89 / 89 & 70 / 60 & NA / NA & 88 / \textbf{89}\\ 
            
            $\text{ResNet-50}_{pgd}$ & 92.36 & 89.08 / 87.55 & 86.09 / 78.62 & 85.09 / 81.04 & 77.14 / 76.35\\ 
            \textbf{Memory-Defense} & 92.36 & \textbf{90.02} / \textbf{90.03} & \textbf{89.99} / \textbf{89.99} & \textbf{89.97} / \textbf{89.80} & \textbf{89.34} / 88.00\\ 
            \bottomrule
            \end{tabular}
        \end{center}
    \end{table*}
    \begin{table*}
        \caption{\label{tab:cifar} Model Performance - \textbf{CIFAR-10} ($L_\infty$ = 2/8/16), where pixel values are clipped in [0,255]} 
        \vspace{-0.5cm}
        \begin{center}
            \begin{tabular}{c c c c c c}
            \midrule
            \textbf{Model} & CLEAN & FGSM & BIM & PGD & CW\\
            \midrule 
            \midrule 
            
            $\text{ResNet-50}_{ben}$ & \textbf{92.78} & 73.46 / 72.19 / 70.40 & 69.69 / 58.50 / 51.49 & 69.60 / 57.76 / 53.39 & 47.63 / 45\\ 
            
            PixelDefend & 90 & 81 / 70 / 67 & 81 / 70 / 56 & NA / NA / NA & 81 / 80\\ 
            
            ComDefend & 91 & 86 / 84 / 83 & 78 / 41 / 34 & NA / NA / NA & 89 / 87\\ 
            
            $\text{ResNet-50}_{pgd}$ & 92.00 & 86.32 / 84.04 / 81.20 & 83.87 / 72.40 / 59.73 & 83.83 / 71.74 / 63.67 & 49.72 / 48.31\\ 
            \textbf{Memory-Defense} & 92.01 & \textbf{86.95} / \textbf{86.94} / \textbf{86.93} & \textbf{86.97} / \textbf{86.97} / \textbf{86.94} & \textbf{87.00} / \textbf{86.11} / \textbf{85.01} & \textbf{89.10} / \textbf{88.33}\\ 
            \bottomrule
            \end{tabular}
        \end{center}
    \end{table*}

\subsection{Memory Selector}

    We choose to implement our memory selector via the popular ResNet-50 classifier \cite{resnet} that returns a probability for each class. The highest probability selects the memory slot.
    
    (Presumably other architectures would also work.) As for the input to the memory, an image and label input is necessary for training and a testing scenario would only contain the input example. Hence, classifier-A plays a significant role in the reconstruction process through memory slots. Memory slots are the critical component for the reconstruction process, as the expectation is to separate each class's latent space as much as possible as shown in Figure \ref{fig:figure3}, which can lead to better reconstruction even if the image holds complex noise. The memory slots are allocated equally to each class and assigned sequentially, i.e., if N represents the total memory size for all the classes then $m_{i-j}, m_{j+1-k}, ..., m_{k+1, l}$, such that $m \in M$. Choosing the wrong memory slot will reconstruct the image with a larger reconstruction error which indicates a wrong choice of memory slot. 
    
    Although classes are mutually exclusive, the majority of classes happen to have a lot of image contents in common. This also results in a smaller reconstruction error from a wrongly selected memory slot. To counter this problem, we pass the $\Bar{w}$ through a headshrinker process similar to Gong et al., \cite{gong2019memorizing} that leaves minimal memory contents for reconstruction. To further increase the gap between the right and the wrong memory slot, we hard-shrink the weight matrix before masking. Without hard-shrinkage, memory attention becomes capable of reconstructing anomalous and adversarial images as accurately as a benign class. 
    
    For the training of classifier-A as well as stacked autoencoder, we use data augmentation techniques traditionally used for DNN classifiers, including random cropping, horizontal flipping, vertical flipping, and a non-traditional method for classifier-A by Zong et al., \cite{zhong2020random} that randomly erases a sequential set of pixels.
    
    We train classifier-A first with the benign examples and use it for reconstructing the adversarial examples that are later used for training with adversarial examples only over PGD attack. Training with PGD has demonstrated robustness to a large variety of $\mathcal{L}_\infty$ distance metric attacks \cite{madry2018towards}. 

\subsection{Proposed Objective Loss}
\label{sec:loss}

    We present a custom loss function to minimize the reconstruction error from the target memory slot ($\mathcal{L}_{\text{target}}$) and at the same time penalizing all the non-target memory slots ($\mathcal{L}_{\text{non-target}}$) by maximizing their reconstruction error as shown below in Equation~\ref{eq:13}. 
    
    The resultant overall loss would normally be bounded by 0. Since we penalize the non-target loss, the subtraction makes the resulting loss unbounded. Hence, we penalize the non-target loss by a constant factor $\beta$ = 1e-4 and and softly clip the non-target reconstruction loss through a sigmoid activation function. 
    The target and non-target loss terms contain a Mean Squared Error (MSE) loss and an Entropy loss. The MSE loss for $\mathcal{L}_{\text{target}}$ (in Equation~\ref{eq:11a}) reconstructs the target image keeping it within the manifold boundary of the target class whereas MSE loss for $\mathcal{L}_{\text{non-target}}$ (in Equation~\ref{eq:11b}) attempts to push the reconstruction far away from the manifold considering the input to the non-target as a suspicious image.
    \[
        \label{eq:13} \small{\mathcal{L}(x,y,w,\hat{w},\hat{x},x',\hat{y}) = \mathcal{L}_{\text{target}}-\beta \cdot \frac{1}{1+e^{- (\frac{1}{N-1} \cdot \mathcal{L}_{\text{non-target}})}}}
        \tag{13}
    \]
    
    \begin{subequations}
        \small{
        \begin{gather}
            \label{eq:11a} \mathcal{L}_{\text{target}} = \frac{1}{B}\sum_{i=1}^{B}(x-\hat{x})^{2}_{\text{y}=\hat{y}}+\alpha\cdot\sum_{i=1}^{B}-w_{i}\cdot log(w_{i})\\
            \label{eq:11b} \mathcal{L}_{\text{non-target}} = 
            \frac{1}{B}\sum_{i=1}^{B}(x-x')^{2}_{\text{y} \neq \hat{y}}+\alpha\cdot\sum_{i=1}^{B}-\hat{w_{i}}\cdot log(\hat{w_{i}})
        \end{gather}
        }
    \end{subequations}

    The second term in  Equations~{\ref{eq:11a} and \ref{eq:11b}} contains the Entropy loss term. We minimize the attention weight $w$ using the entropy loss of the attention weight vector. Here, weight matrix $w_{i}$ have all the entries as positive in the range [0,1] and sum up to 1.
    
    During inference, we calculate the mean squared error (MSE) loss to measure the reconstruction error of the input example. 

    \begin{table*}
        \caption{Models’ classification performance against clean and adversarial data - \textbf{F-MNIST, CIFAR-10} ($\epsilon=0.3$), where pixel values are clipped in [0,255]} \vspace{-0.5cm}
        \label{tab:defense-vae}
        \begin{center}
            \begin{tabular}{ccccccc}
            \toprule
            \multirow{2}{*}{\textbf{Model}} & \multicolumn{3}{c}{\textbf{F-MNIST}} & \multicolumn{3}{c}{\textbf{CIFAR-10}} \\
                                   & CLEAN   & FGSM    & CW      & CLEAN    & FGSM    & CW      \\\midrule\midrule
            Defense-VAE-REC        & 90.85   & 89.03   & 87.47   & 86.52    & 48.52   & 45.91   \\
            Defence-VAE-E2E        & 90.85   & \textbf{91.02}   & 88.54   & 86.52    & 50.72   & 49.44   \\
            \textbf{Memory-Defense}         & \textbf{92.36}   & 89.40   & \textbf{92.36}   & \textbf{92.01}    & \textbf{86.64}   & \textbf{91.00}  \\\bottomrule
            \end{tabular}
        \end{center}
    \end{table*}
    
\section{Experiments}

    \paragraph{Datasets and Experimental Setup.}
    We evaluate our model extensively over the two most popular benchmark datasets; Fashion-MNIST \cite{fmnist}, and CIFAR-10 \cite{cifar10} datasets. Both datasets comprise images of ten mutually exclusive classes. For Fashion-MNIST, the neural-net architecture has three convolutional layers, each for encoder and decoder. CIFAR-10 has four convolutional layers, each for encoder and decoder, as CIFAR10 is a slightly more complex dataset with three-channel inputs. Table \ref{tab:arch} shows the entire architecture of the model. Each layer in both the datasets has a batch-normalization (BN) layer for reducing the covariance-shift and stability of the model and a Leaky Rectified Linear Unit (LeakyReLU) with a negative slope of 1/4 except for the last layer. The bottleneck layer of the encoder uses sigmoid as the activation layer. We use Gaussian noise with ($\phi$=0.0001) for training the autoencoder on F-MNIST whereas, Gaussian noise and L2-regularization for the training on CIFAR-10. The learning rate is 1e-4 for both models with Adam optimizer with betas in the range [0.9, 0.999] and eps=1e-8.
    
    Classifier-A is trained on benign and successful adversarial examples from PGD attack with eps=0.03 for both datasets. We split the fashion-MNIST dataset into 60,000, and 10,000 for training, and testing. For CIFAR-10, we use 50,000 training, and 10,000 testing images. For BIM attack, we use number of iterations=10, epsilon iteration=0.05, clipping=[0,1]. For PGD attack, we use Cross-Entropy loss with reduction=sum as the loss function, iterations=40, epsilon iteration=0.0075, and clipping=[0,1]. Lastly, for CW attack we use the default values as in Cleverhans \cite{cleverhans} with iterations=1000 and learning rate=5e-3, and binary search steps=5.

    \begin{table}[t]
    \caption{\label{tab:arch} Model Architectures}
    \vspace{-0.5cm}
    \small
        \center
        \renewcommand{\arraystretch}{0.95}
        \begin{tabular}{cc}
        \toprule
        \multicolumn{2}{c}{\textbf{Fashion-MNIST}} \\
        \toprule
        \multicolumn{1}{c}{\textit{Encoder}} & \multirow{1}{*}{\textit{F x K x S}} \\ \hline 
            Conv.2D + LeakyReLU &  16 x (1,1) x 2  \\ \hline
            Conv.2D + LeakyReLU &  32 x (3,3) x 2  \\ \hline
            Conv.2D + Sigmoid   &  64 x (3,3) x 2  \\ \hline
        \multicolumn{1}{c}{\textit{Decoder}} & \multirow{1}{*}{\textit{F x K x S}}\\ \hline 
            Conv.Trans.2D + LeakyReLU &  64 x (3,3) x 2  \\ \hline
            Conv.Trans.2D + LeakyReLU &  32 x (3,3) x 2  \\ \hline
            Conv.Trans.2D &  16 x (3,3) x 2  \\
        \end{tabular}
        \quad
        \begin{tabular}{cc}
        \toprule
        \multicolumn{2}{c}{\textbf{CIFAR-10}} \\
        \toprule
        \multicolumn{1}{c}{\textit{Encoder}} & \multirow{1}{*}{\textit{F x K x S}} \\ \hline
            Conv.2D + LeakyReLU & 64 x 2 x 1  \\ \hline
            Conv.2D + LeakyReLU & 128 x 2 x 1  \\ \hline
            Conv.2D + LeakyReLU & 128 x 2 x 1  \\ \hline
            Conv.2D + Sigmoid   & 256 x 2 x 1 \\ \hline
        \multicolumn{1}{c}{\textit{Decoder}} & \multirow{1}{*}{\textit{F x K x S}} \\ \hline
            Conv.Trans.2D + LeakyReLU & 256 x 2 x 1 \\ \hline
            Conv.Trans.2D + LeakyReLU & 128 x 2 x 1 \\ \hline
            Conv.Trans.2D + LeakyReLU & 128 x 2 x 1  \\ \hline
            Conv.Trans.2D & 3 x 2 x 1  \\
        \end{tabular}
        \quad
        \begin{tabular}{cc} \toprule
        \multicolumn{2}{c}{\textbf{Target Classifier}} \\
        \toprule
            Conv.2D + MaxPool2d + ReLU & 6 x 5 x 1   \\ \hline
            Conv.2D + MaxPool2d + ReLU & 16 x 5 x 1  \\ \hline
            Linear  + ReLU & 256 x 256 \\ \hline
            Linear  + ReLU & 256 x 128 \\ \hline
            Linear  & 128 x 10 \\ \hline
        \end{tabular}
    \end{table}
    
    During testing, the system only has access to the input image. To reconstruct the image with a memory slot, we use a classifier and later validate the reconstruction in the pipeline; this makes classifier-A a necessary component in the workflow. One way Memory Defense helps in defense is by reconstruction using relevant memory content makes the task easier for a simple classifier (classifier-B) to get the right prediction.
    
    Adversarial attacks are broadly classified into targeted and non-targeted attacks. We only use non-targeted attacks (in which attacks only aim to force the DNN models to misclassify and do not focus on specific targets) for a direct comparison to the baselines.

    \begin{table}[]
        \begin{center}
        \caption{Ablation Study based on \textbf{F-MNIST dataset}} \vspace{-0.20cm}
        \label{tab:ablation}
        \begin{adjustbox}{width=\columnwidth,center}
            \begin{tabular}{ccc}
            \toprule 
            \textbf{Models}                   & CLEAN & FGSM (25) \\ \midrule\midrule
            Target Classifier                 & 87.62 & 53.64     \\ 
            Autoencoder w/o Memory            & 89.70 & 68.69     \\
            $\text{Memory Selector}_{pgd}$    & 92.36 & 87.55     \\ 
            Memory Defense           & \textbf{92.36} & \textbf{90.03}     \\ \bottomrule
            \end{tabular}
        \end{adjustbox}
        \end{center}
    \end{table}

    \paragraph{Results.}
    
    Tables \ref{tab:fmnist} \& \ref{tab:cifar} present the experimental results of the overall pipeline for the Fashion-MNIST and CIFAR-10 datasets respectively. We compare our testing results with  PixelDefend \cite{song2018pixeldefend} and ComDefend \cite{JiaWCF19}. The results for for the baselines are reported directly from the authors' papers, and PGD is not provided by any of them. We consider PGD as a very powerful attack and it can simply cause failure in a network without defense. For both datasets, Memory Defense results in a significant improvement.
    
    In order to understand the robustness of Memory Defense, we test our model on $L_\infty$ and $L_2$ distance metrics and generate attacks using our model. The fashion-MNIST results show that we achieve \~{}90\% for FGSM attack, an increase in accuracy of 1\%. The performance of our model significantly increases over BIM attack where ComDefend performs at 70\% and 60\%, our model is able to achieve an improvement of 19.99\% and 29\% on BIM attacks. In comparison with PixelDefend we significantly improve by 4.99\% and 6.99\% over BIM attacks. Row 4 shows the classifier-A, performing alone but trained on PGD attack, experiences performance decreases as the perturbation increases on FGSM, BIM, and PGD. Similarly, in Tab.~\ref{tab:cifar} we show experiments on the CIFAR-10 dataset, where we compare our approach over with four baselines. Our approach significantly outperforms all four attacks. In Tab.~\ref{tab:defense-vae} we present results using a very different value for epsilon to compare with the baselines Defense-VAE-REC and Defense-VAE-E2E \cite{DBLP:conf/pkdd/LiJ19} on both F-MNIST and CIFAR-10 data. We report their best accuracy out of the various sets of classifiers (A-D) for each attack and compare it with our approach. We improve by 2\% on clean data and 4\% on CW attack for F-MNIST, whereas our Memory Defense provides significantly higher classification accuracy on the CIFAR-10 dataset.
    The performance also helps us justify that our model is not sensitive to increase in perturbations as it is still able to detect the relevant memory contents for reconstruction using the decoder. This can be seen in Tab.~\ref{tab:fmnist} \& \ref{tab:cifar} where the epsilon ranges from .007-.3.
    
    The overall model's processing time for an average of 10 images for autoencoder without and with memory is 0.00171196 sec and 0.00619208 sec. Adding memory adds an overhead time of \~{}0.004 sec.
    We perform all testing on an Nvidia GeForce RTX 2080 Ti.
    
    \paragraph{Ablation Study.}
        
    Table \ref{tab:ablation} presents an ablation study to understand the effectiveness of each component's contribution and further understand its capability. We first try the prediction capability of the target classifier without an autoencoder and without a memory attention module. The benign accuracy shows the base capability of the TC is almost 88\% but it severely drops on attacks. Adding a vanilla autoencoder slightly decreases the performance for clean images and majorly decreases when attacked because a simple autoencoder lacks the capability for disentangling the feature space properly. Lastly, memory masking significantly improves performance; memory is helping the latent representation in learning the right memory contents of the image and at the same time making it easier for the target classifier to make the prediction. The masking is helping the model to converge the stack faster and sparsity supports the stability as well as prevent the stack from over-fitting.
    
    \paragraph{Memory Size.}
    
\begin{figure}
        \center
        \includegraphics[scale=0.4]{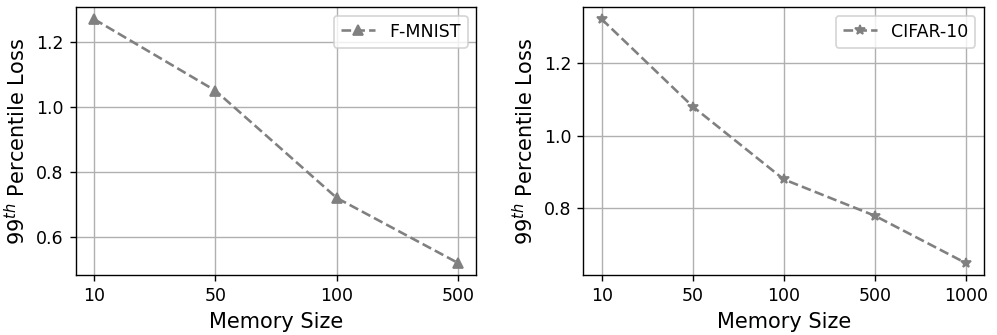}
        \includegraphics[scale=0.195]{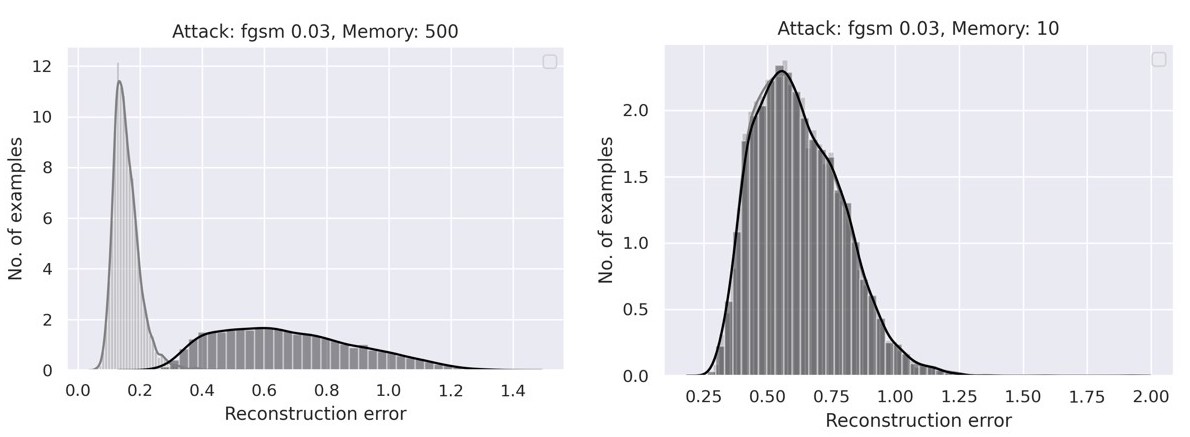}
        \caption{\textit{Memory size setting for each dataset based on $99^{th}$ percentile test loss of the defense model.}}
        \label{fig:figure4} 
    \end{figure}

    Figure \ref{fig:figure4} presents the effect of memory size using two plots. The first pair of plots shows the $99^{th}$ percentile reconstruction error for both the datasets. Based on the above observation, we hypothesized that increasing the dataset's complexity would produce almost similar reconstruction errors. It is evident by the first pair of graphs that increasing the memory size for complex datasets such as CIFAR-10 would lead to similar reconstruction errors. The second pair of graphs shows that using the appropriate memory dimension helps differentiate the benign images from adversarial examples.
    
\section{Discussion}
    ComDefend is a powerful defense tool, but its test/experimental results are limited to the classifier's robustness and not the ComDefend model itself. Hence, ComDefend is expected to have a much lower defense accuracy from attacks generated using the model. In contrast, the idea of Memory Defense is to learn the essential memory features of an image, making the model less sensitive to attacks. Moreover, the target classifier has minimal layers and the architecture is independent of the complexity of the dataset.
    
    Defense-VAE uses a similar proxy-based prediction strategy, but observing their experiments on MNIST and Fashion-MNIST in Table~1 in the main paper shows a significant decrease in performance from \~{}94.90\% to \~{}79.24\% despite both belonging to a single-channel image category (less-complicated). However, this is not the case for Memory Defense on comparing just Fashion-MNSIT to their Fashion-MNIST performance. Comparing the decrease in accuracy from Fashion-MNIST to CIFAR-10 is also not as huge as Defense-VAE.

\section{Conclusion}

    With an increasing growth of applications using machine learning, it is crucial to understand a model's robustness and improve it against adversarial attacks. In this paper, we present a Memory Defense mechanism for mitigating adversarial attacks on the image classification task. We validated our model workflow against four different popular attack schemes. Memory Defense shows significant improvement over two state-of-the-art methods and is thus a valuable new method for mitigating a variety of attacks.
    
    No existing defense method is completely perfect and a given defense mechanism might show good performance but behave differently for different attacks. A potential future direction could be to assemble two or more defense mechanisms for reaching diversity in the overall system against adversarial attacks. Another logical extension would be to train the Memory Defense for generating a robust prediction and flag label simultaneously.

\bibliographystyle{IEEEtran}
\bibliography{arxiv-v1.bib}

\end{document}